%% file: root.tex
\documentclass[letterpaper, 10 pt, conference]{ieeeconf}

\IEEEoverridecommandlockouts

\usepackage{graphics}
\usepackage{siunitx}
\usepackage{mathtools}
\usepackage{amssymb}
\usepackage{braket}
\usepackage{pdfpages}
\usepackage[hidelinks]{hyperref}

\def\transpose{^\top}

\title{\LARGE \bf
Clarke Transform and Clarke Coordinates --
\\A New Kid on the Block for State Representation of Continuum Robots
}

\author{Reinhard M.~Grassmann and Jessica Burgner-Kahrs%
\thanks{We acknowledge the support of the Natural Sciences and Engineering Research Council of Canada (NSERC), [RGPIN-2019-04846].}
\thanks{All authors are with Continuum Robotics Laboratory, Department of Mathematical and Computational Sciences, University of Toronto, Mississauga, ON L5L 1C6, Canada {\tt\small reinhard.grassmann@utoronto.ca}}
}

\begin{document}

\maketitle

\input{Introduction/introduction}
\input{Methods/methods}
\input{Discussion/discussion}
\input{Conclusion/conclusion}

\addcontentsline{toc}{section}{REFERENCES}
\bibliographystyle{IEEEtran}
\bibliography{IEEEabrv, references}

\newpage


\section*{Supplementary material}

# Clarke Transform in a Nutshell
Cheat Sheet for mathematical modelling of continuum robots with $n$ symmetrical arranged displacement-actuated joints.

## General setup and idea

Schematics of a displacement-actuated robot:

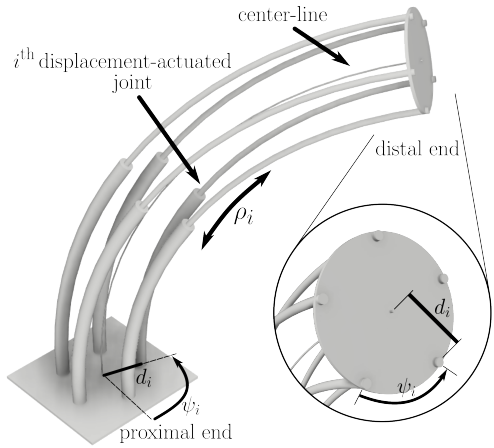

Its sufficient smooth center-line has length $l$. Each of the displacement-actuated joint $\rho_i$ are equal distributed described by $\psi_i = 2\pi(i-1)/n$ and $d_i = d > 0$.

Commutative diagram-like overview:

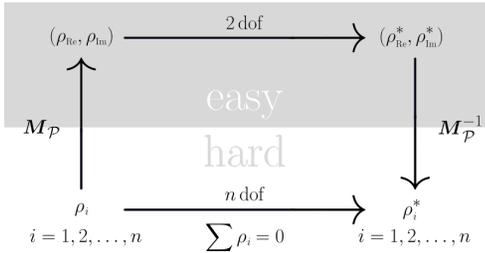

To circumvent the explicit consideration of the displacement constraint, i.e., $\sum \rho_i = 0$, approaches and methods should be considered on the two-dimensional manifold embedded in the $n$-dimensional joint space. For the transform, linear maps $M_\mathcal{P}$ and $M_\mathcal{P}^{-1}$ are used and any output denoted by $(\rho_{\text{Re}}^*, \rho_{\text{Im}}^*)$ can subsequently mapped back leading to $\rho^*$.

## Clarke transform and Clarke coordinates

Generalized Clarke transformation matrix:
$$M_\mathcal{P} = \frac{2}{n}\begin{bmatrix} \cos(0) & \cos\left(2\pi\frac{1}{n}\right) & \cdots & \cos\left(2\pi\frac{n-1}{n}\right) \\ \sin(0) & \sin\left(2\pi\frac{1}{n}\right) & \cdots & \sin\left(2\pi\frac{n-1}{n}\right) \end{bmatrix}$$

Generalized inverse Clarke transformation matrix:
$$M_\mathcal{P}^{-1} = \begin{bmatrix} \cos(0) & \sin(0) \\ \cos\left(2\pi\frac{1}{n}\right) & \sin\left(2\pi\frac{1}{n}\right) \\ \vdots & \vdots \\ \cos\left(2\pi\frac{n-1}{n}\right) & \sin\left(2\pi\frac{n-1}{n}\right) \end{bmatrix}$$

Representation of $i^{\text{th}}$ displacement:
$$\rho_i = \rho_{\text{Re}} \cos \psi_i + \rho_{\text{Im}} \sin \psi_i$$

Kirchhoff's rule is the displacement constraint:
$$\sum_{i=1}^{n} \rho_i = 0$$

Set of displacement-actuated joint values:
$$\boldsymbol{\rho} = \begin{bmatrix} \rho_1 & \rho_2 & \cdots & \rho_{n-1} & \rho_n \end{bmatrix}^\top$$

Clarke coordinates as vector:
$$\overline{\boldsymbol{\rho}} = \begin{bmatrix} \rho_{\text{Re}} & \rho_{\text{Im}} \end{bmatrix}^\top$$

The Clarke transform of $\boldsymbol{\rho}$ and $\overline{\boldsymbol{\rho}}$, respectively:
$$\overline{\boldsymbol{\rho}} = M_\mathcal{P} \boldsymbol{\rho} \quad \text{and} \quad \boldsymbol{\rho} = M_\mathcal{P}^{-1} \overline{\boldsymbol{\rho}}$$

Two-dimensional Manifold embedded in the joint space:
$$\mathbf{Q} = \{\, (\rho_1, \cdots, \rho_n) \in \mathbb{R}^n \mid \forall i \in [1, n] \subset \mathbb{N} : \\ \rho_i = \rho_{\text{Re}} \cos \psi_i + \rho_{\text{Im}} \sin \psi_i \wedge \\ (\rho_{\text{Re}}, \rho_{\text{Im}}) \in \mathbb{R}^2 \,\}$$

## Properties

Linearity of $M_\mathcal{P}$ and $M_\mathcal{P}^{-1}$:
$$M_\mathcal{P} \sum_i \boldsymbol{\rho}_i = \sum_i M_\mathcal{P} \boldsymbol{\rho}_i = \sum_i \overline{\boldsymbol{\rho}}_i$$
$$M_\mathcal{P}^{-1} \sum_i \overline{\boldsymbol{\rho}}_i = \sum_i M_\mathcal{P}^{-1} \overline{\boldsymbol{\rho}}_i = \sum_i \boldsymbol{\rho}_i$$

$M_\mathcal{P}^{-1}$ is the right-inverse of $M_\mathcal{P}$:
$$M_\mathcal{P} M_\mathcal{P}^{-1} = I_{2 \times 2} \quad \text{and} \quad M_\mathcal{P}^{-1} M_\mathcal{P} \neq I_{n \times n}$$

Toeplitz matrix:
$$M_\mathcal{P}^{-1} M_\mathcal{P} = \left( \frac{2}{n} \cos\left(2\pi \frac{i-j}{n}\right) \right)_{i,j} \in \mathbb{R}^{n \times n}$$

Toeplitz matrix is an idempotent matrix:
$$\left( M_\mathcal{P}^{-1} M_\mathcal{P} \right)^k = M_\mathcal{P}^{-1} M_\mathcal{P} \quad \text{for } k > 0$$

Toeplitz matrix is singular:
$$\det M_\mathcal{P}^{-1} M_\mathcal{P} = 0$$

Transpose:
$$M_\mathcal{P}^\top = \frac{2}{n} M_\mathcal{P}^{-1}$$

Vanishing bias term:
$$M_\mathcal{P} \mathbf{1}_{1 \times n} = \mathbf{0}_{1 \times 2},$$
where $\mathbf{1}_{1 \times n}$ has ones everywhere and $\mathbf{0}_{1 \times 2}$ has zeros everywhere.

Selecting a mode:
$$M_\mathcal{P} \mathbb{1}_{1 \times n}^{(k)} = \begin{bmatrix} \cos\left(2\pi \frac{k-1}{n}\right) & \sin\left(2\pi \frac{k-1}{n}\right) \end{bmatrix}^\top,$$
where $\mathbb{1}_{1 \times n}^{(k)}$ is a one-hot vector defined by the $k^{\text{th}}$ element to be a one, whereas the all other elements are zero.

## Properties cont'd

The sum of squares:
$$\rho_{\text{Re}}^2 + \rho_{\text{Im}}^2 = \overline{\boldsymbol{\rho}}^\top \overline{\boldsymbol{\rho}} = \boldsymbol{\rho}^\top \boldsymbol{M}_\mathcal{P}^\top \boldsymbol{M}_\mathcal{P} \boldsymbol{\rho}$$

Scaled magnitude:
$$\overline{\boldsymbol{\rho}}^\top \overline{\boldsymbol{\rho}} = \frac{2}{n} \boldsymbol{\rho}^\top \boldsymbol{\rho}$$

Transform a unit circle:
$$\boldsymbol{M}_\mathcal{P}^{-1} \begin{bmatrix} \cos(\alpha) & \sin(\alpha) \end{bmatrix}^\top = (\cos(\psi_i - \alpha))_{i,1}$$

Visual aid and geometric interpretation:

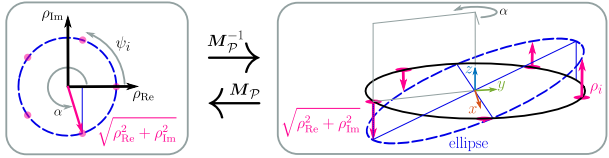

Tracing the tip of all possible displacement parameterized by $\psi \in [0, 2\pi)$ creates an ellipse. Its semi-major and semi-minor axes are $\sqrt{d^2 + \rho_{\text{Re}}^2 + \rho_{\text{Im}}^2}$ and $d$ length, respectively. The maximum displacement achievable coincides with the angle $\alpha$.

Normalized displacement-actuated joints:
$$\widehat{\boldsymbol{\rho}} = \frac{\boldsymbol{\rho}}{\sqrt{\boldsymbol{\rho}^\top \boldsymbol{\rho}}}$$

To avoid singularities, add a sufficient small $\epsilon > 0$ with $\epsilon^2 \approx 0$ to the magnitude, i.e., $\sqrt{\boldsymbol{\rho}^\top \boldsymbol{\rho}} + \epsilon$

## Trigonometric Identities

Useful trigonometric identities for $\psi_i = \frac{2\pi}{n}(i-1)$:

$$\sum_{i=1}^n \sin(\psi_i) = 0, \quad \sum_{i=1}^n \sin^2(\psi_i) = \frac{n}{2},$$
$$\sum_{i=1}^n \cos(\psi_i) = 0, \quad \sum_{i=1}^n \cos^2(\psi_i) = \frac{n}{2},$$
$$\text{and} \quad \sum_{i=1}^n \sin(\psi_i)\cos(\psi_i) = 0$$

## Arc space and virtual displacement

Physical interpretation of the Clarke coordinates:

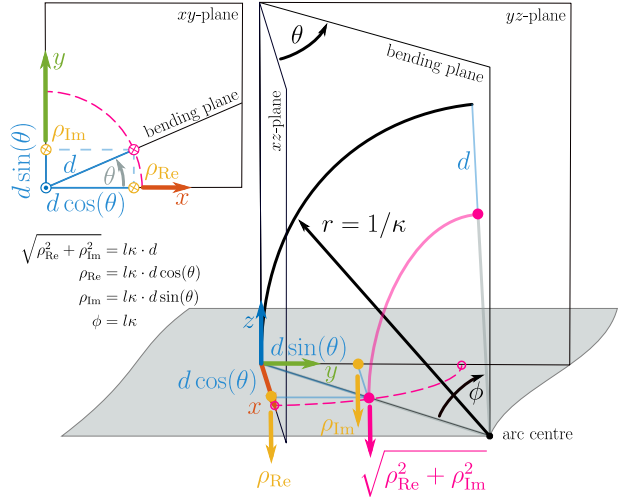

The magenta line lies within the bending plane. The length difference to the arc length is the virtual displacement. The yellow arrows are the projected virtual displacements and lie within the respective projected plane corresponding to $xz$-plane and $yz$-plane of the base.

Relation to arc space:
$$\boldsymbol{M}_\mathcal{P} \boldsymbol{\rho} = \begin{bmatrix} \rho_{\text{Re}} & \rho_{\text{Im}} \end{bmatrix}^\top = ld \begin{bmatrix} \kappa \cos(\theta) & \kappa \sin(\theta) \end{bmatrix}^\top$$

The design parameters, i.e., segment length $l$ and joint location $(\phi_i, d_i)$, are removed, i.e.,

$$\begin{bmatrix} \kappa \cos(\theta) \\ \kappa \sin(\theta) \end{bmatrix} = \underbrace{1/l}_{\text{removes } l} \underbrace{\boldsymbol{M}_\mathcal{P}}_{\text{removes } \psi_i} \underbrace{\text{diag}(1/d_i)}_{\text{removes } d_i} \boldsymbol{\rho}.$$

For its inverse, the design parameters are added, i.e.,

$$\boldsymbol{\rho} = \underbrace{l}_{\text{adds } l} \underbrace{\text{diag}(d_i)}_{\text{adds } d_i} \underbrace{\boldsymbol{M}_\mathcal{P}^{-1}}_{\text{adds } \psi_i} \begin{bmatrix} \kappa \cos(\theta) \\ \kappa \sin(\theta) \end{bmatrix}.$$

For both formulation, the assumption $d_i = d$ has been removed.

## Application to tendon-driven continuum robot

Schematics of a tendon-driven continuum robot:

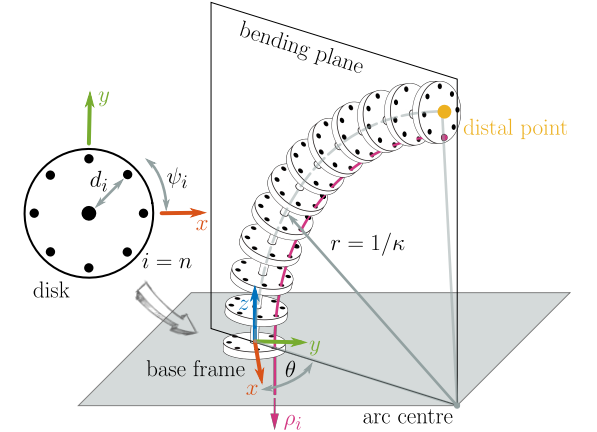

Application to tendon control:

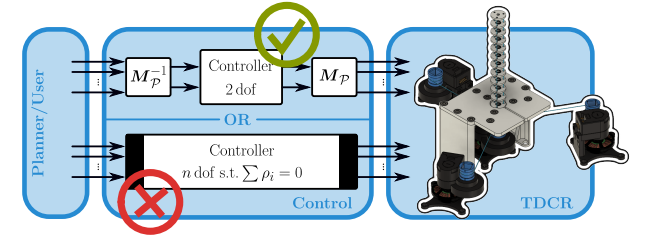

Reducing the design effort and the computational cost by choosing a suitable space to formulate the control problem.

Control schema:

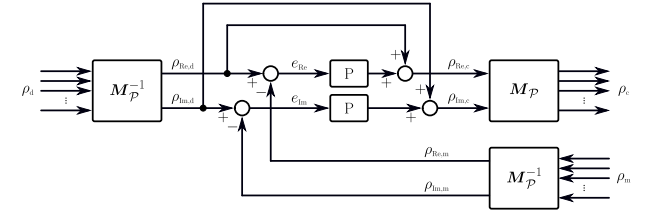

Displacement-control of $n$ displacements using Clarke transform. Both proportional feedback controller with precompensation are sandwich by $\boldsymbol{M}_\mathcal{P}$ and $\boldsymbol{M}_\mathcal{P}^{-1}$.



\end{document}

%% file: Introduction/introduction.tex
\section{Motivation}
A tendon-driven continuum robot (TDCR) operates under constraints intrinsic to its mechanical design.
For almost all TDCRs, a segment is actuated by three or four tendons using a parallel tendon routing. 
For both cases, methods to account for the constraints are known, \textit{i.e.}, the tendon displacement across all tendons should sum up to zero.
However, for the general case, \textit{i.e.}, arbitrary number of tendons, a disentanglement method has yet to be formulated.
Motivated by this unsolved general case, we explored state representations of TDCRs and exploited the underlying two-dimensional manifold.
We found that the Clarke transformation (also known as $\alpha\beta\gamma$ transformation), a mathematical transformation used in vector control (also known as field-oriented control), can be generalized to address this problem.

In this extended abstract, we build on the Clarke transformation used to simplify the analysis of three-phase circuits and derive the generalized Clarke transformation for arbitrary number of phases, which can then be applied to TDCRs with arbitrary number of tendons.
We refer to this generalization as the \textit{Clarke transform} and the \textit{Clarke coordinates}.
Furthermore, by investigating the physical meaning of the Clarke coordinates, a connection to robots with constant curvature characteristics is established.

%% file: Methods/methods.tex
\section{Clarke Transform in a Nutshell}

\begin{figure}[t]
    \vspace{0.75em}
    \centering
    \includegraphics[width=\columnwidth]{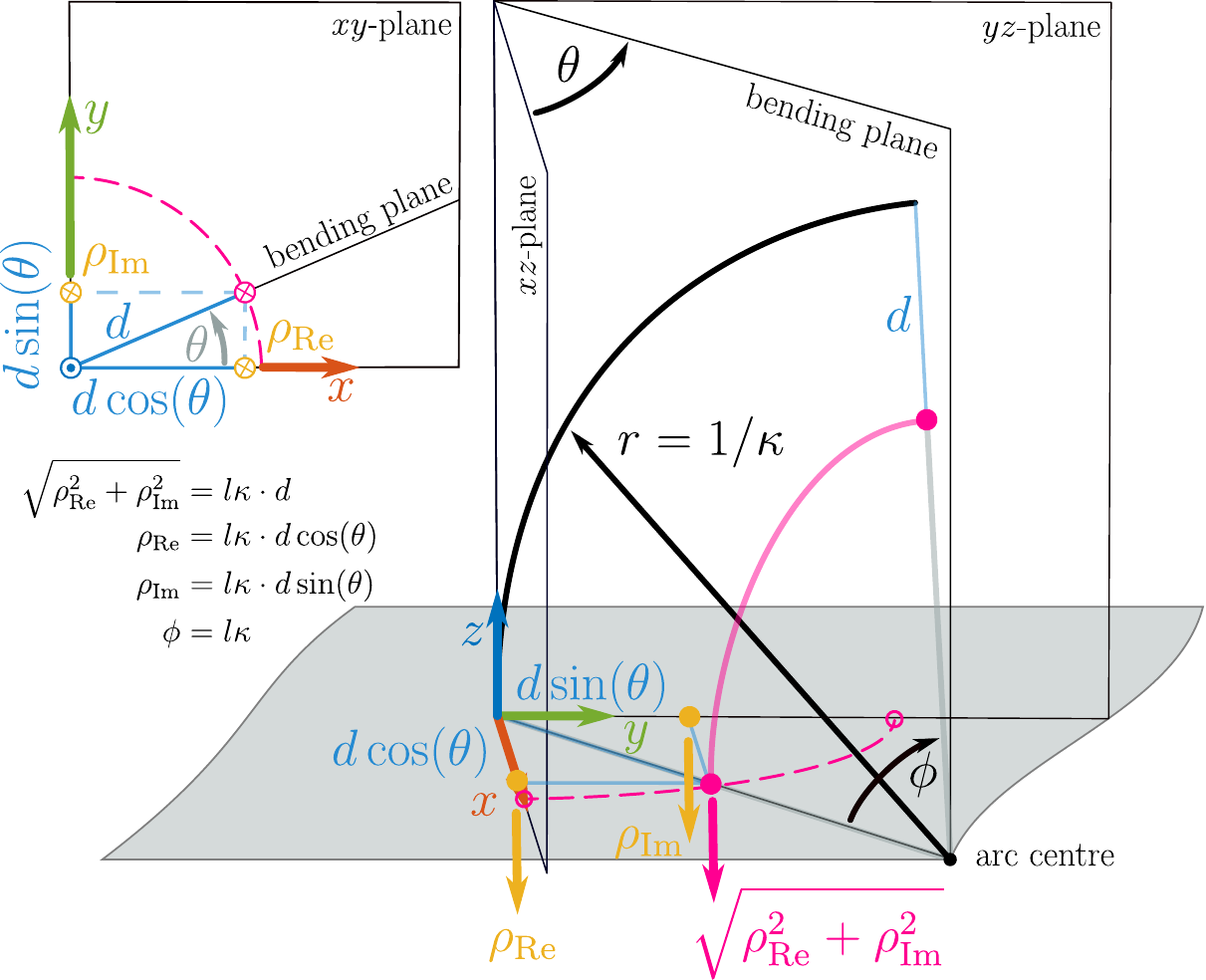}
    \vspace{-2em}
    \caption{
        Physical interpretation of the Clarke coordinates.
        The magenta line lies within the bending plane.
        The length difference to the arc length is the virtual displacement.
        The yellow arrows are the projected virtual displacements and lie within the respective projected plane corresponding to $xz$-plane and $yz$-plane of the base.
    }
    \label{fig:virtual_displacement}
    \vspace{-1.5em}
\end{figure}

The Clarke transform is mathematically consistent as well as physically interpretable.
Here, we give a quick-and-dirty introduction to Clarke transform.
Fig.~\ref{fig:virtual_displacement} provides visual aid by highlighting its geometric interpretation.

The \textit{Clarke transform} utilized a \textit{generalized Clarke transformation} matrix $\boldsymbol{M}_\mathcal{P} \in \mathbb{R}^{2 \times n}$ being defined as
\begin{flalign}
    %
    \small 
    \!\!
	\boldsymbol{M}_\mathcal{P}
    \!=\!
    \dfrac{2}{n}\!
	\begin{bmatrix}
		\cos\left(0\right) & \cos\left(2\pi\dfrac{1}{n}\right) & \cdots & \cos\left(2\pi\dfrac{n-1}{n}\right)\\[0.85em]
		\sin\left(0\right) & \sin\left(2\pi\dfrac{1}{n}\right) & \cdots & \sin\left(2\pi\dfrac{n-1}{n}\right)
	\end{bmatrix}
	\label{eq:M_P}
\end{flalign}
Its inverse is the right-inverse of  $\boldsymbol{M}_\mathcal{P}$ defined as $\boldsymbol{M}_\mathcal{P}^{-1} = \left(n/2\right)\boldsymbol{M}_\mathcal{P}\transpose$.

For a TDCR, we can define a joint space representation $\rho_i = \rho_\mathrm{Re}\cos{\psi_i} + \rho_\mathrm{Im}\sin{\psi_i}$
describing the displacement of the $i\textsuperscript{th}$ tendon located at angle $\psi_i = 2\pi(i - 1)/n$ and distance $d_i = d > 0$ using polar coordinates.
The two free parameters of $\rho_i$, \textit{i.e.}, $\rho_\mathrm{Re} \in \mathbb{R}$ and $\rho_\mathrm{Im} \in \mathbb{R}$, are the local coordinates of the \SI{2}{dof} manifold embedded in the $n\,\SI{}{dof}$ joint space for $n$ tendon displacements $\rho_i$.
We call them \textit{Clarke coordinates} in honor of Edith Clarke.
The set $\boldsymbol{Q} = \Set{\boldsymbol{\rho} \in \mathbb{R}^n | \rho_i = \rho_\mathrm{Re}\cos{\psi_i} + \rho_\mathrm{Im}\sin{\psi_i}\,\wedge\,\sum \rho_i = 0}$
is a subset of $\mathbb{R}^n$ defining the joint space of all possible tendon displacements, where $\boldsymbol{\rho} = \left[\rho_1,\ \rho_2,\ \cdots,\ \rho_n\right]\transpose$.

As a consequence of the formulation, the Clarke transform of $\boldsymbol{\rho}$ results in the Clarke coordinates.
More importantly, a linear relationship to the commonly used arc parameters \cite{WebsterJones_IJRR_2010} can be found (see Fig.~\ref{fig:virtual_displacement}):
\begin{align}
    \boldsymbol{M}_\mathcal{P}
    \boldsymbol{\rho} 
    =
    \begin{bmatrix} 
        \rho_\mathrm{Re}&
        \rho_\mathrm{Im} 
    \end{bmatrix}\transpose
    = 
    ld
    \begin{bmatrix}
        \kappa\cos\left(\theta\right) & 
        \kappa\sin\left(\theta\right)
    \end{bmatrix}\transpose
    ,
    \label{eq:forward_mapping}
\end{align}
with arc length $l$, survature $\kappa$, and bending plane angle $\theta$.

\begin{figure}[t]
    \centering
    \vspace*{0.75em}
    \includegraphics[width=0.9\columnwidth]{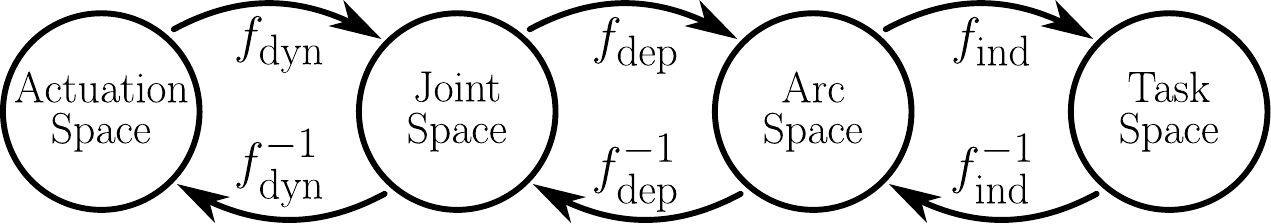}
    \vspace*{-0.5em}
    \caption{
    Spaces and their mappings.
    To map between the actuator, \textit{e.g.}, motor angles to joint values, \textit{e.g.}, tendon displacements, $f_\text{dyn}$ and $f_\text{dyn}^{-1}$ are used.
    The robot-dependent mapping is denoted by $f_\text{dep}$, whereas robot-independent mapping is denoted by $f_\text{ind}$.
    Their inverses are $f_\text{dep}^{-1}$ and $f_\text{ind}^{-1}$.
    }
    \vspace*{-1.75em}
    \label{fig:spaces_spaces}
\end{figure}

Applying the Clarke transform and Clarke coordinates alongside trigonometric relationships, we revisited the kinematics of TDCR.
In fact, Eq.~\eqref{eq:forward_mapping} is a closed-form solution to the forward robot-dependent mapping for the general case of TDCR, i.e. $n$ tendons per segment instead of 3 or 4.
Furthermore, it is inherently providing a closed-form inverse robot-dependent mapping.
Both, $f_\text{dep}$ described by Eq.~\eqref{eq:forward_mapping} and its inverse $f_\text{dep}^{-1}$ are the previously missing transformation between the joint and arc space, see Fig.~\ref{fig:spaces_spaces}.

In fact, it can be shown, that the different variations of improved state representations proposed in recent works by Qu \textit{et al.} \cite{QuLau_et_al_ROBIO_2016}, Della Santina \textit{et al.} \cite{DellaSantinaBicchiRus_RAL_2020}, Cao \textit{et al.} \cite{CaoXie_et_al_JMR_2022}, and Dian \textit{et al.} \cite{DianGuo_et_al_Access_2022}, can be derived by using the Clarke transform.
Consequently, the Clarke transform unifies existing state representations and, more importantly, generalizes them to $n$ displacement-actuated joints providing a robot-type agnostic approach and enabling the advantages gained by all improved state representations.

%% file: Discussion/discussion.tex
\section{Potential Impact and Novel Directions}

While many interesting directions can be pursued, we focus on a broader implication inherent to most of them.
The constant curvature assumption \cite{WebsterJones_IJRR_2010} is a widely used approach for a variety of continuum and soft robots.
In fact, the tendon displacement constraint, \textit{i.e.}, $\sum \rho_i = 0$, and the Clarke transform are relevant for all constant curvature robots with symmetrically arranged displacement-actuated joints $\rho_i$, \textit{e.g.}, tendons, push-pull rods, and bellows.

A general mapping between spaces could be 
\begin{align}
    \boldsymbol{\rho}
    \rightleftharpoons 
    \left(\rho_\mathrm{Re}, \rho_\mathrm{Im}\right)
    \rightleftharpoons 
    \left(\kappa\cos\left(\theta\right), \kappa\sin\left(\theta\right)\right)
    \rightleftharpoons 
    \left(\boldsymbol{p}, \xi\right)
    ,
    \nonumber
\end{align}
where $\boldsymbol{\rho}$ can be any displacement-actuated joint regardless of the robot type.
This chain of transformations delineates a geometric mapping from the quaternion/vector-pair $\left(\boldsymbol{p}, \xi\right)$, to the arc parameters $\left(\kappa\cos\left(\theta\right), \kappa\sin\left(\theta\right)\right)$ and subsequently to Clarke coordinates $\left(\rho_\mathrm{Re}, \rho_\mathrm{Im}\right)$.
This mapping exploits the geometric properties inherent to each component, effectively utilizing their structures for enhanced computational efficiency and clarity in algorithms. This approach is aligned with the current trend in robotics research, which emphasizes geometric methodologies into robot learning, optimization, and control.

\begin{figure}
    \vspace{0.5em}
    \centering
    \includegraphics[width=0.85\columnwidth]{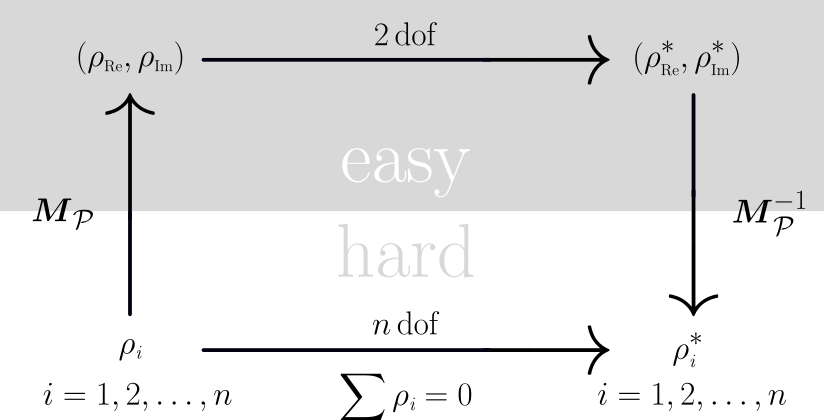}
    \vspace*{-1em}
    \caption{
    Commutative diagram-like overview of the joint space disentanglement usage.
    To circumvent the explicit consideration of the actuator  constraints, \textit{i.e.}, $\sum \rho_i = 0$, methods can act on the \SI{2}{dof} manifold embedded in the $n\,$\SI{}{dof} joint space.
    For the transform, linear maps $\boldsymbol{M}_\mathcal{P}$ and $\boldsymbol{M}_\mathcal{P}^{-1}$ are used and any output denoted by $\left(\rho_\text{Re}^*, \rho_\text{Im}^*\right)$ can subsequently mapped back leading to $\rho^*$.
    }
   \vspace*{-1.5em}
    \label{fig:commutative_diagram}
\end{figure}

Adopting Clarke coordinates as a standard state representation could standardize fundamental methodologies, making research tools like task planners, control schemes, and machine learning algorithms reusable across different research communities.
Methods can be developed based on Clarke coordinates instead of a case-by-case base for different types of constant curvature robots.
The Clarke transform is a key approach that provides the possibility to also simplify the modeling aspect, which can lead to conceptually simpler and more efficient approaches thanks to the dimensionality reduction.
Inspired by depictions of the well-known Laplace transform and Fourier transform, a high-level illustration of the workflow is shown in Fig.~\ref{fig:commutative_diagram}.

Ultimately, we believe that the Clarke transform and Clarke coordinates will enable the effective control of robots which leverage more actuators per segment, \textit{e.g.}, $7$ or $11$ tendons. 
While mechanically more complex, the merit of many more actuators than the common three or four should be exploited. 
It is our hypothesis that increased load capacity, better shape conformation, increased stability, and variable stiffness could be achieved.

To avoid stagnation and the risk of becoming too inwardly focused, we follow the suggestion by Hawkes \textit{et al.} \cite{HawkesMajidiTolley_SR_2021} and push beyond approaches that only contribute to our continuum robotics community.
The Clarke transform has the potential to push the boundaries and provide a higher level of contribution \cite{HawkesMajidiTolley_SR_2021} for future publications.
Its linearity, simplicity, interpretability, generalizability to known improved representation, and deep relation to the constant curvature model render the Clarke transform as an important stepping stone for modelling, control, design, and beyond that eventually can lead to many novel methods and mechanical designs that are quantitatively superior to the state-of-the-art methods.

%% file: Conclusion/conclusion.tex
\section{Conclusion}

In this extended abstract, we briefly introduce the Clarke transform and touch on its potential impact on continuum robotics.
We are convinced that the Clarke transform will have a similar impact on continuum robots and soft robots as the constant curvature assumption.
Through follow-up work, time will tell how the Clarke Transform can facilitate advanced and unified modelling, control, and design approaches.